\documentclass{article}

\usepackage{arxiv}

\usepackage[utf8]{inputenc} 
\usepackage[T1]{fontenc}    
\usepackage{hyperref}       
\usepackage{url}            
\usepackage{booktabs}       
\usepackage{amsfonts}       
\usepackage{nicefrac}       
\usepackage{microtype}      
\usepackage{lipsum}
\usepackage{graphicx}
\usepackage{xcolor}
\graphicspath{ {./images/} }

\title{Neuro-symbolic model for cantilever beams damage detection}

\author{
Darian Onchiș \\
 Department of Computer Science\\
 West University of Timisoara\\
 Timisoara, Romania \\
 \texttt{darian.onchis@e-uvt.ro} \\
  \And
Gilbert-Rainer Gillich \\
 Department of Engineering Science\\
 Babes Bolyai-University\\
 Cluj-Napoca, Romania \\
 \texttt{gilbert.gillich@ubbcluj.ro} \\
 \And
Eduard Hogea \\
 Department of Computer Science\\
 West University of Timisoara\\
 Timisoara, Romania \\
 \texttt{eduard.hogea00@e-uvt.ro} \\
 \And
Cristian Tufiși \\
 Department of Engineering Science\\
 Babes Bolyai-University\\
 Cluj-Napoca, Romania \\
 \texttt{cristian.tufisi@ubbcluj.ro} \\
}

\begin{document}
\maketitle
\begin{abstract}
In the last decade, damage detection approaches swiftly changed from advanced signal processing methods to machine learning and especially deep learning models, to accurately and non-intrusively estimate the state of the beam structures. But as the deep learning models reached their peak performances, also their limitations in applicability and vulnerabilities were observed.  One of the most important reason for the lack of trustworthiness in operational conditions is the absence of intrinsic explainability of the deep learning system, due to the encoding of the knowledge in tensor values and without the inclusion of logical constraints. In this paper, we propose a neuro-symbolic model for the detection of damages in cantilever beams based on a novel cognitive architecture in which we join the processing power of convolutional networks with the interactive control offered by queries realized through the inclusion of real logic directly into the model. The hybrid discriminative model is introduced under the name Logic Convolutional Neural Regressor and it is tested on a dataset of values of the relative natural frequency shifts of cantilever beams derived from an original mathematical relation. While the obtained results preserve all the predictive capabilities of deep learning models, the usage of three distances as predicates for satisfiability, makes the system more trustworthy and scalable for practical applications. Extensive numerical and laboratory experiments were performed, and they all demonstrated the superiority of the hybrid approach, which can open a new path for solving the damage detection problem.
\end{abstract}


\section{Introduction}


Continuous monitoring of engineering structures (CMES) has become an increasingly common practice in the attempt to ensure their safe operation \cite{singh2022realtime}. The monitoring process involves a series of actions such as data acquisition, post-processing, and analysis. Monitoring leads, in this way, to the early identification of defects that have occurred in equipment \cite{baker2009towards} or structural elements \cite{sevieri2020dynamic}. By permanently knowing the state of the engineering structure, maintenance can be performed at the most appropriate time with lower costs and increased security of the equipment and structures in operation. This type of maintenance, known as condition-based maintenance (CMB), is becoming increasingly popular in the Industry 4.0 era. The implementation of CBM is favored by the development of cheap and reliable sensors and advanced data mining algorithms \cite{raheja2006data}. It is worth mentioning that CBM can be easily integrated within the manufacturing processes.

If we refer to structures, nondestructive control usually implies the assessment of cracks because these are the most common structural damage. Assessing cracks means finding the location and the severity, and sometimes the type of the crack \cite{wavelet10}. The assessment of damage can be made by local methods such as visual inspection, liquid penetrant testing, magnetic particle testing, ultrasonic testing, acoustic emissions, infrared thermography, and radiographic testing. A comprehensive review is made in \cite{Nou1}. These methods have the disadvantage of evaluating the integrity of structures in a local area to which access is frequently required.
On the other hand, global methods do not require access to the damaged area, since these assess the global health of the structure. The global methods can be divided into vibration-based \cite{Nou2, Nou3, onur2021review} and static methods \cite{Nou4,Isac10}. Vibration-based methods (VBM) use data obtained from multiple vibration modes in contrast to static ones that use only displacements that are similar to the data obtained for the first mode. Therefore, VBMs can better identify the location and severity of damages.

Considering the above-mentioned, we propose a in-here a detection method based on the analysis of the dynamic response of structures to impulsive or continuous excitations. The method makes use of the relationship existing between the modal parameters’ changes and the position and geometry of a crack \cite{bowman2013can}. In section 4, we show how a dataset containing numerous crack scenarios and the resulting frequency changes is constructed. An insight into the method is also given in this section. Because of the dimension and the complexity of the dataset, finding the crack scenario producing a certain set of measured frequency changes is difficult. A good solution to overcome this problem is to use artificial intelligence (AI) techniques. Deep learning may be the answer for one such issue \cite{onchis2}, but we also must consider the scenarios where it may fail. 

\section{Related works}

Given the numerous researchers and papers that review deep learning strengths and weaknesses for CMES, concerns such as the issues of adversarial attacks, the opaqueness of the model and the extensive need for large datasets and high computational costs, made us lean towards implementing a different kind of system.
These are just some issues deep learning faces, with the first one maybe playing a future important role in the next possible AI winter, considering the current overhype and unprecedented increase in funding. Papers such as \cite{heaven2019deep}, \cite{akhtar2018threat} study the effects of adversarial attacks and how simple their implementation is. Adversarial attacks can range from noise, to blocking on part of an image, to just adjusting a single pixel and the model can be easily fooled. Examples such as 3D printed turtles being classified as rifles, a bell pepper as a strainer, a bus as an ostrich and a stop sign being classified as a speed limiter are just some of those. The main problems of deep learning have been greatly identified in \cite{marcus2018deeplearning}, where the author provides an explanation of the biggest concerns it faces. Among those, the ones aforementioned are also presented. A suggestion in this mentioned paper about dealing with those concerns is to supplement deep learning with other techniques, and suggests the need for hybrid models.

As a solution to incorporate data and logic and to overcome Deep Learning deficiencies, in \cite{badreddine2016logictensornetworks} the Logic Tensor Networks (LTN) were proposed. These neuro-symbolic models combine neural networks and first-order logic language. With fuzzy logic, this framework provides reasoning over data, and it can be used to design a regression model. LTNs reduce the learning problem of a given formula by optimizing its satisfiability \cite{synasc22}. The network will try to optimize the groundings to bring the truth of the formula close to 1.

Some papers where similar hybrid approaches may be useful are \cite{lerer2016learning} and \cite{bates2015humans} where the Deep Learning models used for intuitive physics would benefit from having prior knowledge about dynamics. Similar approaches that use Logic Tensor Networks have been used to classify images \cite{donadello2017logictensor}, with defined relational knowledge for robustness), sentiment analysis \cite{huang2022logictensor}, and for using prior knowledge for transfer learning \cite{bowman2013can}.

But for the accurate processing of datasets with relative frequency shifts (RFS) that characterize the state of the beams, 1D convolutional neural networks provide superior results than feed-forward fully connected neural architectures. We base our statement in the translation invariance that Conv1D layers provide and their ability to learn local patterns within the input sequence, but also in the numerical experiments performed in this paper.Therefore, we propose in here a novel neuro-symbolic architecture under the name Logic Convolutional Neural Regressor (LCNR) that joins the advantages of convolutional neural networks (CNN) with the intrinsic explainability offered through the incorporation of Real Logic (RL) into the model design. RL is described by a first-order language that contains constants, variables, functions, predicates, and logical connectives and quantifiers. The way LCNR uses Real Logic is by transforming the provided formulas into TensorFlow computational graphs. Those formulas can then be used for querying and learning, and make it possible to have interactive accuracy and deductive reasoning over data.

\section{Main Contributions}

Our primary contribution to the field of computer applications for industry is the design and implementation of a novel Logical Convolutional Neural Regressor (LCNR) neuro-symbolic system, which performs non-invasive damage detection on a cantilever beam RFS dataset. This hybrid cognitive system combines deep learning techniques with symbolic reasoning, resulting in an intrinsically explainable model that provides insight into its decision-making process. The system employs convolutional layers to handle complex spatial patterns in the data and utilizes three distinct predicates grounded in Real Logic to apply constraints during the training of our regression model. This innovative approach allows for improved interpretability and robustness in comparison to traditional regression methods, as well as the incorporation of logical relationships between features, which can be particularly valuable in cases where the dataset is more complex.

By integrating neuro-symbolic networks, our model demonstrates how logical reasoning can be effectively combined with deep learning techniques, enabling the encoding of domain-specific knowledge and the extraction of meaningful relationships within the data. This approach has the potential to enhance the performance and explainability of neural networks in various industrial applications, particularly when dealing with intricate datasets. To validate the effectiveness of our model,  cross-validation was utilized on a shuffled dataset, with our analysis incorporating multiple folds for each predicate. This rigorous evaluation process ensured the reliability of the model's performance across various data subsets.

The trained model was used to predict and compare the damage position on a real cantilever beam, demonstrating its practical applicability in real-world scenarios. Real Logic queries were employed to ensure trustworthiness in operation conditions, providing additional confidence in the model's ability to accurately assess structural damage.

In summary, our main contribution lies in the development of a highly interpretable and explainable LCNR neuro-symbolic system, capable of effectively detecting damage in cantilever beams and offering potential for further advancements in the field of structural health monitoring and other industrial applications that require complex feature relationships and domain-specific knowledge.
\section{Construction of the RFS dataset}
This study concerns the identification of damage in cantilever beams, which can be subject to ideal or non-ideal clamping at one end. Obviously, the second end of the beam is free.  For these beams, we create a dataset containing a series of scenarios regarding the clamping condition, the crack position, and the crack depth. A schematic of the beam highlighting the clamping condition and the crack parameters is presented in Figure 1 is affected by imperfect clamping conditions by using the relative frequency shift method. The algorithm for generating the training datasets was developed in previous research.

\begin{figure}[!h]
    \centering
    \includegraphics[width = 1\textwidth]{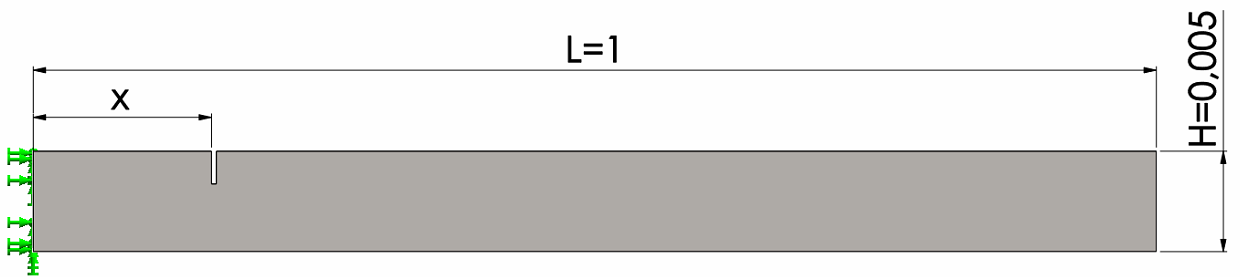}
    \caption{Schematic cantilever beam}
    \label{fig:my_label}
\end{figure}

The mathematical relation relies on the natural frequencies of the structure in undamaged state, the modal curvature of the first eight weak-axis vibration modes, and the severity of the transverse cracks, as established in \cite{gillich2019robust}. The equation \ref{first eq} is given by:
\begin{equation}
\label{first eq}
    f_{i-D}(x,a) = f_{i-U}\left \{ 1-\gamma (a)\ [\bar{\phi}_i''(x)]^2\ \right \}
\end{equation}
where we noted:
\begin{itemize}
    \item $f_{i-U}$ the natural frequency for the undamaged beam, 
    \item $f_{i-D}$ the natural frequency for the damaged beam
    \item $x$ crack position  
    \item $a$ crack depth 
    \item $\gamma(a)$ damage severity 
    \item $\left [ \phi_i''(x) \right ] ^2$ squared normalized modal curvature for the i-th mode of transverse vibration
\end{itemize}
From equation \ref{first eq} the RFS values are deduced with the following relation \ref{second eq}:
\begin{equation}
    \label{second eq}
    \Delta \bar{f}_{i-D} (x,a) = \frac{f_{i-U} - f_{i-D}(x-a)} { {f_{i-U}}} = \gamma(a)[\bar{\phi}_i''(x)]^2
\end{equation}
The severity $\gamma(a)$ of closed and open transverse cracks can be determined by using the method presented in \cite{gillich2019robust}. The squared normalized modal curvature $\left [ \phi_i''(x) \right ] ^2$, for a cantilever is determined with the procedure described in another work by Gillich and Praisach in \cite{gillich2014modal}. The performance of the methods is confirmed by two independent studies \cite{Nou5, Nou6}.

In addition to the transverse crack, the method also considers the possibility of imperfect fastening of the structure, simulating the real-life scenario where loosening of joints may occur. To consider this possibility, following the approach proposed in \cite{mituletu2019method}, relation \ref{second eq} becomes \ref{third eq}:
\begin{equation}
    \label{third eq}
    \Delta \bar{f}_{i-D}(0, a_1, x_2, a_2) = \gamma_1 (a_1) + \gamma_2 (a_2)[\bar{\phi}_i''(x_2)]^2
\end{equation}
Where:
\begin{itemize}
    \item $\gamma_1 (a_1)$ is the severity of the clamping condition
    \item $\gamma_2 (a_2)$ is the severity of the transverse crack
    
\end{itemize}
By using relation \ref{third eq} we have generated the training data considering a closed transverse crack of depth starting from 4\% to 64\% relative to the beam thickness. 


Also, the training data consists of four scenarios of imperfect boundary conditions, thus permitting a small rotation at the fixed end \cite{Nou7}. Instead of a massless spring \cite{Nou8}, the rotation is possible due to a crack that has the depth from 10\% to 20\% of the beam thickness. It resulted in a total number of 36573 possible damage scenarios, from which we use 70\% for training and 30\% for testing and validating.

The generated dataset is available on the Mendeley repository \cite{tufisi2023damage}. After the model is trained, we generate test datasets by means of FEM simulations and experimental laboratory measurements for several damage scenarios.

\section{Logic Convolutional Neural Regressor}
\begin{figure}[!ht]
    \centering
    \includegraphics[width = \textwidth]{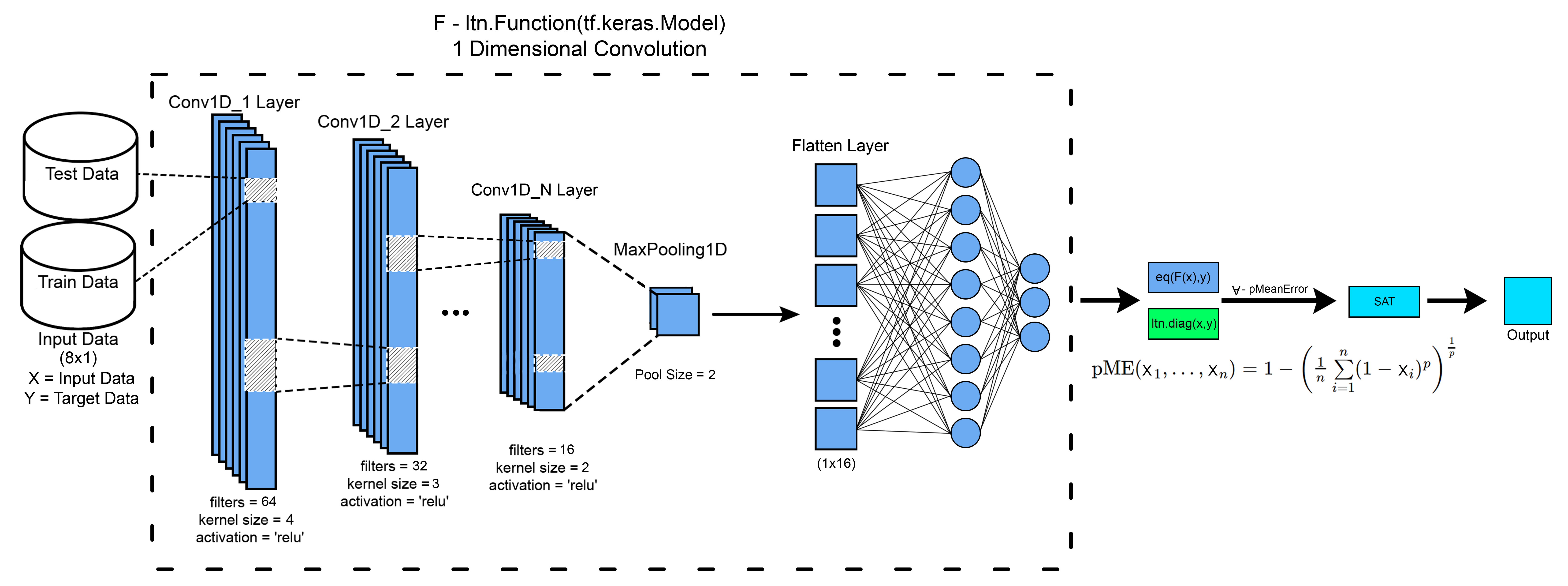}
    \caption{Schematic LCNR model.}
    \label{fig:LCNR}
\end{figure}

The hybrid system in Figure \ref{fig:LCNR} is composed of a deep convolutional neural network and a predicate grounded in Real Logic used to make a regression on beam damage assessment data. The regression problem was solved by using a function that took the argument of a sequential model with multiple Conv1D layers, e.g., 1D Convolution. For our proposed model, data is reshaped and batched into tensors. A function F, as described in Real Logic can be any operation supported by TensorFlow. Considering the benefits of convolutional layers compared to fully connected ones, a CNN network was used. Furthermore GPUs accelerations were implemented to speed up the computational load \cite{face}.

The axioms for this model involve the application of an aggregation operator ($\forall$ - pMeanError) to calculate the distance/similarity between the output of function F and the target data. The diagonal quantification of input and target data, achieved by using lcnr.diag(x, y), allows for statements concerning specific input-output pairs, such as the i-th instance of both individuals.

The learning phase of our model aims to maximize the satisfiability of the proposed formula by optimizing the groundings:
$\forall(lcnr.diag(x,y), eq(Fx,y))$.

In this formula, the $Forall (\forall)$ quantifier iterates over each input-output pair created by lcnr.diag(x, y), and the eq predicate calculates the similarity between the model's output F(x) and the target y for each pair. The pMeanError function is then used to aggregate these individual similarities, and the model's objective is to maximize the overall satisfiability with respect to the similarities of all input-output pairs.

This implementation aims to utilize the supervised technique to predict the crack location in any of the proposed damage scenarios. In our experiments, predicates for Euclidean, Manhattan, and Minkowski of order 1 and 2 (generalization of Manhattan and Euclidean distances) distance/similarity were defined and used to constrain the parameters of the function. During each epoch, the satisfaction levels of the Knowledge Base for both train and test inputs were constantly monitored, alongside the model's accuracy.

Given the goal of predicting the numerical value of the crack position, the Root Mean Square Error (RMSE) was chosen as the accuracy measurement for comparing predicted values with actual ones. However, it is important to note that the primary objective of our model during training was to maximize the satisfiability value of our axioms. Satisfiability, a concept from fuzzy logic, ranges between [0,1] and is analogous to a loss function in Machine or Deep Learning.

Our model uses the RFS from the data set as input and predicts the crack position as in Figure 1. In the last training epochs, the model reached an accuracy of ~0.01-0.03 for our validation data. As such, we considered it was ready for the real-world beam damage assessment scenario. For that and to emphasize that a large data set may not be always available, K-fold cross-validation was used and for each fold, a plot of residuals was made. All the data and results are available in the before mentioned Mendeley repository.

\begin{figure}[!htbp]
    \centering
    \includegraphics[width = 0.4\textwidth]{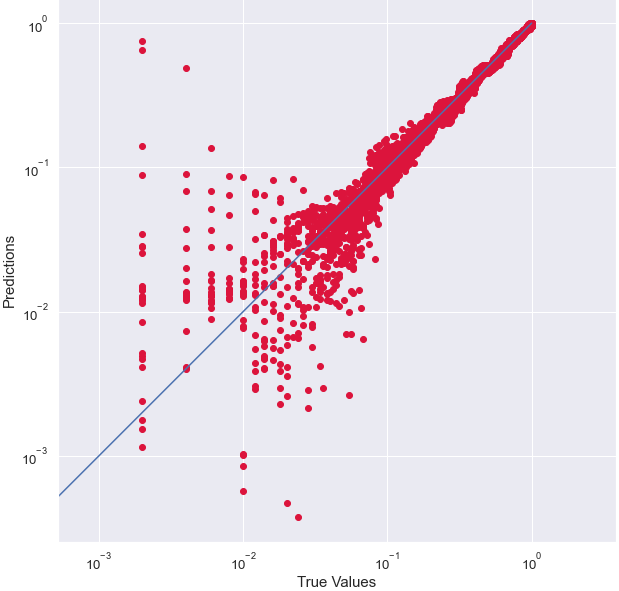}
    \caption{Relationship between the predicted and actual values. The blue line represents a perfect regression.}
    \label{fig:plot LCNR}
\end{figure}

Additional experiments were made and added to the paper repository, with testing on how the model behaves when only a smaller part of the training data is available (10\%, 20\%, 30\%… 90\%). The content for those percentages of data used in training was randomly chosen from the whole data, and the same was done for the validation data. The batch of tensors fed to the neural network, constraining function, the shape of the neural network and the metrics in measurements were kept the same. Results show impressive accuracy even with 10\% of data.
To prove the improvements in accuracy of LCNR, the same data was used as in the first examples and the results of different predicates can be seen in the following section.

\section{Testing the algorithm with real-world data}
To evaluate the performance of our system in a practical setting, where noise, variability, and other challenges may be present, we have decided to test it on a new set of data.


\begin{table}[ht]
    \centering
    \caption{ Results of LCNR (with 4 different metrics) on the FEM dataset, the same model but without the added logical part (Conv1D) and a typical DNN regression model with a similar number of parameters. The listed methods represent the predicate used in our system. The standard deviation is listed to assess the performance. }
    \label{tab:methods}
    \begin{tabular}{lc}
        \toprule
        Method & Standard Deviation\\
        \midrule
        Euclidean & 10.8  \\
        Manhattan & 10.7\\
        Mink, $p=2$ & 11.5\\
        Mink, $p=1$ & 12.0\\
        Conv1D, DL & 48.2\\
        DNN & 51.0\\
        \bottomrule
    \end{tabular}
\end{table}

A more in-depth look is also helpful to see how the system performs. As such, the following tables detail the results, with a side-by-side comparison of actual position, predicted one and the absolute difference between them. In all our test settings, cross validation was used and the worst 4 scenarios, with respect to the error, can be seen in the tables below.

\begin{table}[ht]
    \centering
    \caption{Distance/similarity comparison of the worst predicted scenarios.}
    \label{tab:comparison_corrected}
    \begin{tabular}{lccc@{\hskip 10pt}ccc@{\hskip 10pt}ccc}
        \toprule
        & \multicolumn{3}{c}{Euclidean} & \multicolumn{3}{c}{Manhattan} \\
        \cmidrule(lr){2-4} \cmidrule(lr){5-7}
        & Real & Predicted & Error[\%] & Real & Predicted & Error[\%] \\
        \cmidrule(lr){1-4} \cmidrule(lr){5-7}
        & 325 & 376 & 5.1 & 466 & 414 & 5.2 \\
        & 347 & 397 & 5.0 & 489 & 440 & 4.9 \\
        & 414 & 460 & 4.6 & 516 & 470 & 4.6 \\
        & 690 & 735 & 4.5 & 516 & 472 & 4.4 \\
        \cmidrule(lr){1-7}
        & \multicolumn{3}{c}{Minkowski of order 1} & \multicolumn{3}{c}{Minkowski of order 2} \\
        \cmidrule(lr){2-4} \cmidrule(lr){5-7}
        & Real & Predicted & Error[\%] & Real & Predicted & Error[\%] \\
        \cmidrule(lr){1-4} \cmidrule(lr){5-7}
        & 690 & 646 & 4.4 & 360 & 317 & 4.3 \\
        & 173 & 216 & 4.3 & 360 & 402 & 4.2 \\
        & 255 & 294 & 3.9 & 466 & 507 & 4.1 \\
        & 165 & 204 & 3.9 & 796 & 834 & 3.8 \\
        \bottomrule
    \end{tabular}
\end{table}

Changing the predicate and using a different distance metric proved to impact the performance of our system, but the results are relatively similar. This, however, is not true when we compare with a different approach like Deep Neural Network, where the results are worse as seen in both the standard deviation and the qualitative analysis of samples.

\subsection{FEM dataset for testing}

\begin{figure}
    \centering
    \includegraphics[width = \textwidth]{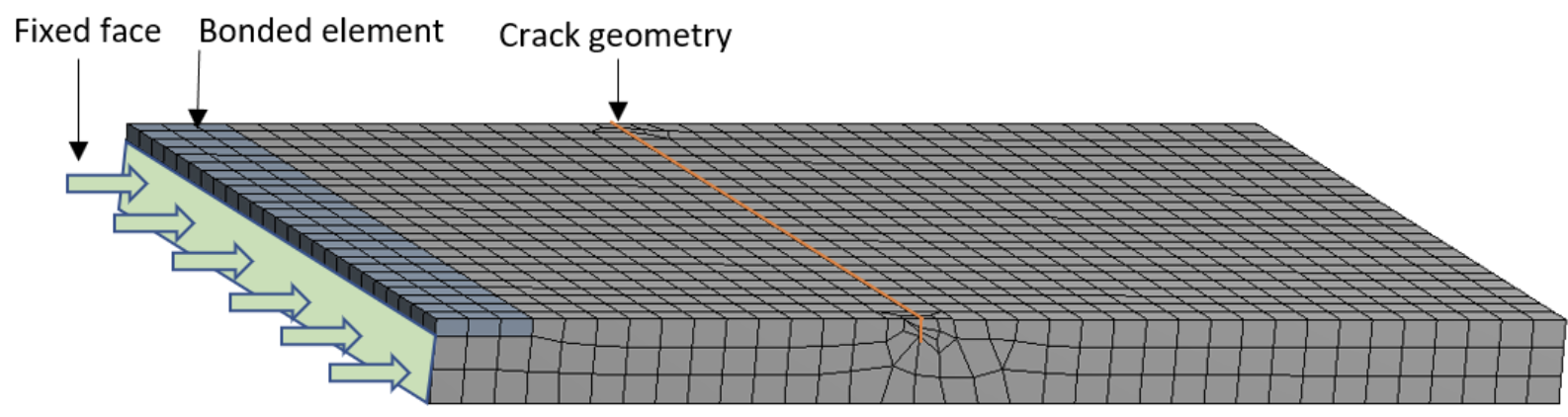}
    \caption{Detail on the fixed end of the model realized in ANSYS with highlighting the crack and the imperfect boundary condition.}
    \label{fig:ANSYS}
\end{figure}

Throughout this study we have considered the structure as an Euler-Bernoulli cantilever beam with the elasticity module $E=2*105$ MPa, mass density $\rho=7850$ kg/m3, length L=1 m, width B=0.05 m and thickness H=0.005 m. The beam’s 3D model is generated and imported in the ANSYS simulation software. The modal study is set by applying the boundary condition, i.e., fixed boundary at the left end of the beam. A fine mesh containing hexahedra elements of 1 mm maximum size was applied. After meshing, the study is run and the natural frequency for the undamaged beam is acquired. After this step, the transverse crack geometry is generated and parametrized in the ANSYS modeler. Several damage scenarios are considered. For simulating the cases with improper clamping, an extra element is defined exactly where the fastened end is, without constraining it. The thickness of this element represents the considered severity $\gamma_1(a_1)$. The FEM simulation setup is presented in Figure \ref{fig:ANSYS}.

After the crack and clamping conditions are defined, we have generated several damage scenarios. The cases are considered having a transverse crack of depth a=1 mm, meaning a severity value $\gamma_2(a_2)=0.0033459$. The crack is located in specific positions considered from the left end, with the values presented in Table \ref{tab:damage_scenarios}.

\begin{table}[ht]
    \centering
    \caption{Defined damage scenarios}
    \label{tab:damage_scenarios}
    \begin{tabular}{c c | c c}
        \toprule
        Damage scenario & Crack position x [mm] & Damage scenario & Crack position x [mm] \\
        \midrule
        1  & 56  & 14 & 466 \\
        2  & 81  & 15 & 489 \\
        3  & 120 & 16 & 516 \\
        4  & 165 & 17 & 560 \\
        5  & 173 & 18 & 660 \\
        6  & 210 & 19 & 687 \\
        7  & 233 & 20 & 690 \\
        8  & 255 & 21 & 760 \\
        9  & 290 & 22 & 796 \\
        10 & 325 & 23 & 820 \\
        11 & 347 & 24 & 896 \\
        12 & 360 & 25 & 906 \\
        13 & 414 & 26 & 946 \\
        \bottomrule
    \end{tabular}
\end{table}
Furthermore, by considering the same crack depth and positions given in Table 3 we have defined weak clamping scenarios, by successively considering two fastening severities $\gamma_1(a_1)=0.0033460$ and $\gamma_1(a_1)=0.0021409$ which represent 24\% and 16\% clamping alteration, resulting in a total number of 78 damage scenarios.
For all cases, the natural frequencies for the first eight weak-axis vibration modes are recorded and the RFS values are obtained by using relation \ref{second eq}.

\subsection{Experimental dataset for testing}
\begin{figure}
    \centering
    \includegraphics[width = \textwidth]{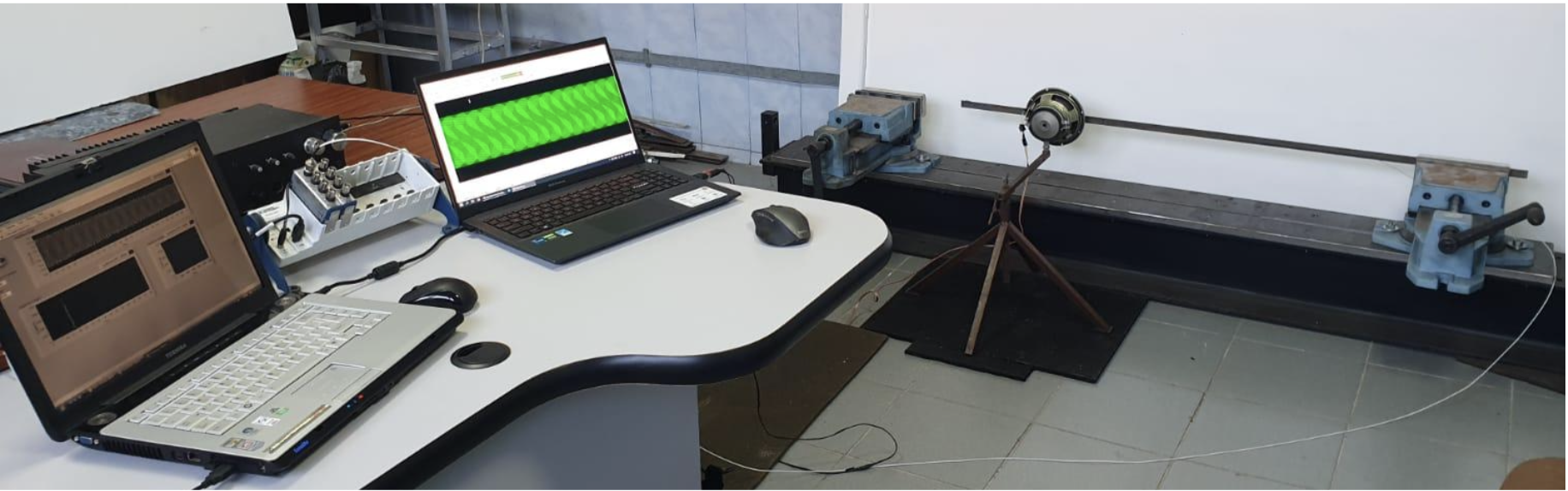}
    \caption{Experimental setup}
    \label{fig:setup}
\end{figure}

In the current subsection we present the methodology for generating the experimental test dataset by measuring the natural frequencies of steel cantilever beam tests in undamaged and in damaged state, also considering a case where a beam is affected by improper clamping.
Every test beam is fixed in a vise, excited using generated sound waves at the desired frequencies and the eigenvalues are recorded through an accelerometer, interface, and special software. Both the excitation and data acquisition procedures are described in detail in the papers \cite{gillich2019robust} \cite{gillich2014modal} \cite{mituletu2019method}.
The experimental setup is presented in Figure \ref{fig:setup}.

The considered crack positions and depths for the cases with perfect clamping are presented in Table \ref{tab:damage_scenarios_2}. For the generated damage depths, we calculate the damage severity ($\gamma$), which is also presented in the before-mentioned table.

\begin{table}[ht]
    \centering
    \caption{Defined damage scenarios}
    \label{tab:damage_scenarios_2}
    \begin{tabular}{lccccc}
        \toprule
        & \multicolumn{5}{c}{Damage scen.} \\
        & Beam 1 & Beam 2 & Beam 3 & Beam 4 & Beam 5 \\
        \midrule
        x [mm]     & 98   & 310  & 569  & 126  & 759  \\
        a [mm]     & 2.5  & 1.25 & 2.5  & 2.5  & 2.5  \\
        $\gamma_2(a)_2$ & 0.0262 & 0.0051 & 0.0262 & 0.0262 & 0.0262 \\
        \bottomrule
    \end{tabular}
\end{table}

\begin{figure}
    \centering
    \includegraphics[width = 0.5\textwidth]{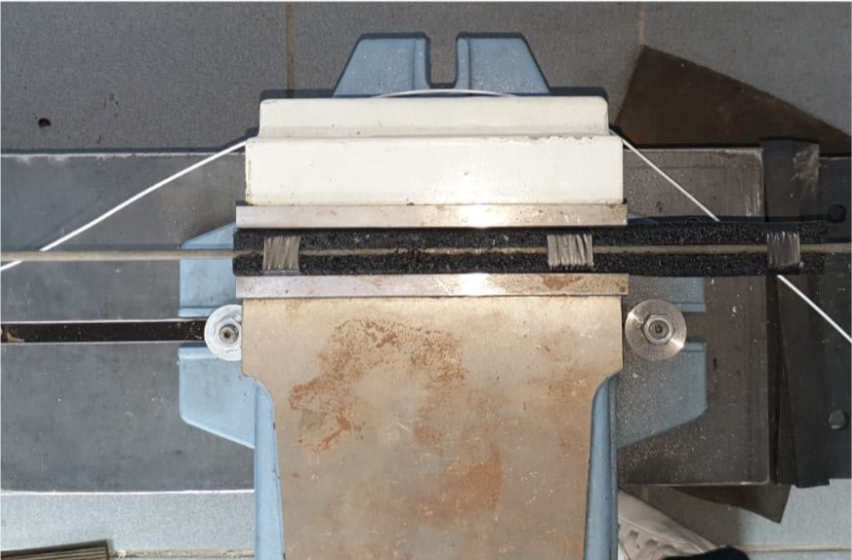}
    \caption{ Imperfect (weak) clamping setup}
    \label{fig:clamp}
\end{figure}

Furthermore, for the weak clamping case, Beam 1 was considered. The setup is presented in Figure \ref{fig:clamp}. 

The obtained RFS values are summarized in Table \ref{tab:rfs_values}. When taking into account the imperfect boundary condition, the test (Beam 1) is fastened by inserting two rubber blocks between the beam and the steel jaws of the vise.

We measure first the vibration response of the beams without the generated damage and estimate the natural frequencies after a procedure described in \cite{mituletu2019method}. Next, we generate damage and repeat the procedure of frequency estimation. Finally, we calculate the RFS values, relation \ref{second eq}.

\begin{table}[ht]
    \centering
    \caption{Obtained RFS values for the test beam}
    \label{tab:rfs_values}
    \begin{tabular}{lcccccc}
        \toprule
        & \multicolumn{6}{c}{Damage scenario} \\
        & Beam 1 & Beam 2 & Beam 3 & Beam 4 & Beam 5 & Beam 1 \\
        \midrule
        Damage location & 98 & 310 & 569 & 126 & 759 & 98 \\
        Crack depth mm & 2.5 & 1.25 & 2.5 & 2.5 & 2.5 & 2.5 \\
        Crack depth severity & 0.026224 & 0.005124 & 0.026224 & 0.026224 & 0.026224 & 0.026224 \\
        Clamping type & Perfect & Perfect & Perfect & Perfect & Perfect & Imperfect \\
        \midrule
        RFS \\
        \midrule
        Mode 1 & 0.020610 & 0.001795 & 0.002382 & 0.023458 & 0.000288 & 0.026955 \\
        Mode 2 & 0.007828 & 0.000660 & 0.017252 & 0.005550 & 0.005461 & 0.013803 \\
        Mode 3 & 0.001629 & 0.002334 & 0.005019 & 0.000064 & 0.017336 & 0.007749 \\
        Mode 4 & 0.000031 & 0.000600 & 0.009109 & 0.002581 & 0.017272 & 0.006322 \\
        Mode 5 & 0.002070 & 0.000542 & 0.011488 & 0.008901 & 0.004422 & 0.008572 \\
        Mode 6 & 0.005964 & 0.002654 & 0.002556 & 0.014021 & 0.000996 & 0.012559 \\
        Mode 7 & 0.009762 & 0.001223 & 0.016603 & 0.014610 & 0.012234 & 0.016089 \\
        Mode 8 & 0.011873 & 0.000146 & 0.000017 & 0.010491 & 0.016377 & 0.017436 \\
        \bottomrule
    \end{tabular}
\end{table}

After the RFS values from Table \ref{tab:rfs_values} are obtained by using relation \ref{second eq}, the accuracy of the developed LCNR is tested for the defined real-life scenarios.

\section{Results and discussions}
The intelligent algorithms are trained using as input the RFS values determined analytically for predicting the position and severity of transverse cracks, even when imperfect clamping is involved. We test the accuracy of the models by using the data obtained through FEM simulations and experimental procedures.
\subsection{Accuracy testing using the FEM results}
We tested the accuracy of the developed model to predict the position of the transverse cracks, by introducing the RFS values obtained for the FEM simulations for all scenarios presented. The obtained results for the scenarios with perfect clamping are presented in Table \ref{tab:results_perfect_clamping}. The error in the following tables refers to the relative discrepancy between the predicted and the actual location of the damage along a bar of 1000mm. It is calculated with the following formula \ref{eq_error}:
\begin{equation}
\label{eq_error}
\mbox{Error [\%]} = \left| \frac{\mbox{Predicted} - \mbox{Real}}{1000} \right| \times 100
\end{equation}

\begin{table}[ht]
    \centering
    \caption{Results obtained for the damage scenarios with perfect clamping}
    \label{tab:results_perfect_clamping}
    \begin{tabular}{clclcclclc}
        \toprule
        & \multicolumn{4}{c}{Damage Scenario} & & \multicolumn{4}{c}{Damage scenario} \\
        \cmidrule{2-5} \cmidrule{7-10}
        & \multicolumn{2}{c}{Crack position x [mm]} & \multicolumn{2}{c}{Error [\%]} & & \multicolumn{2}{c}{Crack position x [mm]} & \multicolumn{2}{c}{Error [\%]} \\
        \cmidrule{2-5} \cmidrule{7-10}
        & Real & Predicted & & & & Real & Predicted & & \\
        \midrule
        1  & 56  & 52.605  & 0.340  & & 14 & 466 & 423.707 & 4.229 & \\
        2  & 81  & 71.092  & 0.991  & & 15 & 489 & 458.925 & 3.007 & \\
        3  & 120 & 96.869  & 2.313  & & 16 & 516 & 460.720 & 5.528 & \\
        4  & 165 & 133.080 & 3.192  & & 17 & 560 & 524.264 & 3.574 & \\
        5  & 173 & 140.818 & 3.218  & & 18 & 660 & 644.529 & 1.547 & \\
        6  & 210 & 183.252 & 2.674  & & 19 & 687 & 675.843 & 1.116 & \\
        7  & 233 & 204.225 & 2.877  & & 20 & 690 & 680.542 & 0.946 & \\
        8  & 255 & 232.158 & 2.284  & & 21 & 760 & 739.184 & 2.082 & \\
        9  & 290 & 281.614 & 0.838  & & 22 & 796 & 766.848 & 2.915 & \\
        10 & 325 & 315.698 & 0.930 & & 23 & 820 & 790.347 & 2.965 & \\
        11 & 347 & 336.876 & 1.012 & & 24 & 896 & 882.467 & 1.353 & \\
        12 & 360 & 349.734 & 1.026 & & 25 & 906 & 899.799 & 0.620 & \\
        13 & 414 & 398.890 & 1.511 & & 26 & 946 & 964.947 & 1.894 & \\
        \bottomrule
    \end{tabular}
\end{table}

\begin{table}[ht]
    \centering
    \caption{Results obtained for the damage scenarios with imperfect clamping}
    \label{tab:results_imperfect_clamping_2}
    \begin{tabular}{clclcclclc}
        \toprule
        & \multicolumn{4}{c}{Damage Scenario} & & \multicolumn{4}{c}{Damage scenario} \\
        \cmidrule{2-5} \cmidrule{7-10}
        & \multicolumn{2}{c}{Crack position x [mm]} & \multicolumn{2}{c}{Error [\%]} & & \multicolumn{2}{c}{Crack position x [mm]} & \multicolumn{2}{c}{Error [\%]} \\
        \cmidrule{2-5} \cmidrule{7-10}
        & Real & Predicted & & & & Real & Predicted & & \\
        \midrule
        27 & 56  & 47.987  & 0.801 & & 40 & 466 & 395.247 & 7.075 & \\
        28 & 81  & 68.215  & 1.279 & & 41 & 489 & 440.911 & 4.809 & \\
        29 & 120 & 100.212 & 1.979 & & 42 & 516 & 468.056 & 4.794 & \\
        30 & 165 & 139.528 & 2.547 & & 43 & 560 & 522.347 & 3.765 & \\
        31 & 173 & 148.006 & 2.499 & & 44 & 660 & 640.515 & 1.949 & \\
        32 & 210 & 184.273 & 2.573 & & 45 & 687 & 669.519 & 1.748 & \\
        33 & 233 & 208.066 & 2.493 & & 46 & 690 & 669.050 & 2.095 & \\
        34 & 255 & 237.016 & 1.798 & & 47 & 760 & 725.672 & 3.433 & \\
        35 & 290 & 280.431 & 0.957 & & 48 & 796 & 763.983 & 3.202 & \\
        36 & 325 & 316.065 & 0.893 & & 49 & 820 & 789.601 & 3.040 & \\
        37 & 347 & 336.730 & 1.027 & & 50 & 896 & 869.310 & 2.669 & \\
        38 & 360 & 336.730 & 2.327 & & 51 & 906 & 883.037 & 2.296 & \\
        39 & 414 & 358.597 & 5.540 & & 52 & 946 & 941.592 & 0.441 & \\
        \bottomrule
    \end{tabular}
\end{table}

\begin{table}[ht]
    \centering
    \caption{Results obtained for the damage scenarios with imperfect clamping}
    \label{tab:results_imperfect_clamping_3}
    \begin{tabular}{clclcclclc}
        \toprule
        & \multicolumn{4}{c}{Damage Scenario} & & \multicolumn{4}{c}{Damage scenario} \\
        \cmidrule{2-5} \cmidrule{7-10}
        & \multicolumn{2}{c}{Crack position x [mm]} & \multicolumn{2}{c}{Error [\%]} & & \multicolumn{2}{c}{Crack position x [mm]} & \multicolumn{2}{c}{Error [\%]} \\
        \cmidrule{2-5} \cmidrule{7-10}
        & Real & Predicted & & & & Real & Predicted & & \\
        \midrule
        53 & 56  & 42.995  & 1.301 & & 66 & 466 & 418.942 & 4.706 & \\
        54 & 81  & 62.682  & 1.832 & & 67 & 489 & 436.844 & 5.216 & \\
        55 & 120 & 97.701  & 2.230 & & 68 & 516 & 468.162 & 4.784 & \\
        56 & 165 & 135.323 & 2.968 & & 69 & 560 & 506.787 & 5.321 & \\
        57 & 173 & 143.408 & 2.959 & & 70 & 660 & 639.915 & 2.008 & \\
        58 & 210 & 178.840 & 3.116 & & 71 & 687 & 665.192 & 2.181 & \\
        59 & 233 & 202.454 & 3.055 & & 72 & 690 & 667.220 & 2.278 & \\
        60 & 255 & 229.125 & 2.587 & & 73 & 760 & 726.848 & 3.315 & \\
        61 & 290 & 273.119 & 1.688 & & 74 & 796 & 764.322 & 3.168 & \\
        62 & 325 & 310.303 & 1.470 & & 75 & 820 & 785.755 & 3.425 & \\
        63 & 347 & 331.844 & 1.516 & & 76 & 896 & 863.200 & 3.280 & \\
        64 & 360 & 348.813 & 1.119 & & 77 & 906 & 875.503 & 3.050 & \\
        65 & 414 & 386.226 & 2.777 & & 78 & 946 & 927.354 & 1.865 & \\
        \bottomrule
    \end{tabular}
\end{table}

The obtained results for the scenarios with 16\% clamping alteration are presented in Table \ref{tab:results_imperfect_clamping_2}. In Table \ref{tab:results_imperfect_clamping_3} the results obtained for the cases with 24\% alteration.

\subsection{Accuracy testing using the experimental measurements}

Upon conducting experimental tests to measure the natural frequencies for the specified damage scenarios, the predictive accuracy of the Logic Convolutional Neural Regressor (LCNR) and its embedded Deep Learning (DL) network in discerning the location of the crack was evaluated separately. The findings of this investigation are delineated in Table \ref{tab:results_imperfect_clamping_4}, where the same inputs have been tested to the two networks. Based on study we have conducted before, the distance/similarity metric used is the one resembling the Euclidean.


\begin{table}[ht]
    \centering
    \caption{Real versus Predicted values for LCNR and the DL Network within it.}
    \label{tab:results_imperfect_clamping_4}
    \begin{tabular}{cccc|cccc}
        \toprule
        \multicolumn{4}{c}{LCNR} & \multicolumn{4}{c}{Conv1D Network (same as the one used in LCNR)} \\
        \midrule
        Real & Predicted & Error[\%] & Type & Real & Predicted & Error[\%] & Clamping type \\
        \midrule
        120.00 & 118.72 & 0.128 & Perfect & 120.00 & 101.62 & 1.838 & Perfect \\
        56.00 & 57.22 & 0.122 & Imperfect & 56.00 & 13.38 & 4.262 & Imperfect \\
        165.00 & 166.06 & 0.106 & Perfect & 165.00 & 168.50 & 0.350 & Perfect \\
        81.00 & 82.04 & 0.104 & Imperfect & 81.00 & 97.70 & 1.670 & Imperfect \\
        \bottomrule
    \end{tabular}
\end{table}

\section{Conclusions}
In this study, we have proposed a neuro-symbolic algorithm, LCNR, for accurately predicting the position and severity of transverse cracks in cantilever steel beams, even under imperfect clamping conditions. Our approach combines the strength of artificial neural networks with symbolic regression. The effectiveness of the LCNR methodology was tested using both FEM simulations and experimental measurements.

Our results demonstrate that LCNR achieves high accuracy in predicting the location of transverse cracks in cantilever steel beams. The algorithm was able to predict damage scenarios with an error of less than 5\% for both FEM and experimental datasets, surpassing the performance of a Deep Neural Network model used as a backbone for LCNR. This validates the effectiveness of the proposed hybrid method and highlights its potential applicability in real-world scenarios where precise damage prediction is vital. 

\section{Acknowledgments}

The artificial intelligence part including the novel LCRN neuro-symbolic deep learning architecture was proposed and implemented by Darian Onchis and Eduard Hogea. The mechanical engineering part including the dataset acquisition and the experiments was realized by Gilbert-Rainer Gillich and Cristian Tufisi.

\bibliographystyle{unsrt}  
\bibliography{references}  


\newpage
\appendix

\section{Annex 1}
\begin{figure}[!ht]
    \centering
    \includegraphics[width = 0.5\textwidth]{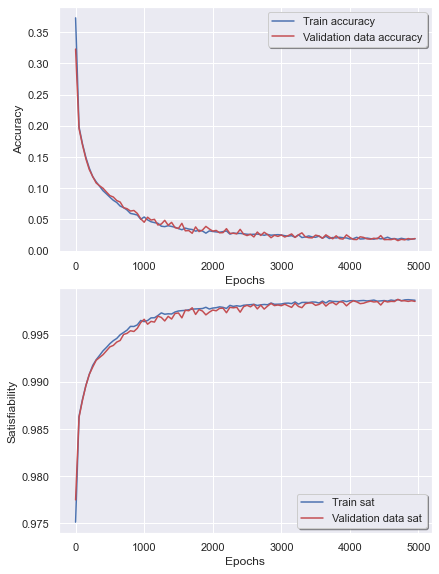}
    \caption{Evolution of accuracy and satisfiability with the Euclidean distance/similarity predicate}
    \label{fig:annex}
\end{figure}

\end{document}